\documentclass[conference]{IEEEtran}
\IEEEoverridecommandlockouts
\usepackage{cite}
\usepackage{amsmath,amssymb,amsfonts}
\usepackage{algorithmic}
\usepackage{graphicx}
\usepackage{textcomp}
\usepackage{xcolor}
\usepackage[caption=false]{subfig}
\usepackage{multirow}
\usepackage{epsfig}

\def\BibTeX{{\rm B\kern-.05em{\sc i\kern-.025em b}\kern-.08em
    T\kern-.1667em\lower.7ex\hbox{E}\kern-.125emX}}

\makeatletter
\def\ps@IEEEtitlepagestyle{%
	\def\@oddfoot{\mycopyrightnotice}%
	\def\@evenfoot{}%
}
\def\mycopyrightnotice{%
	{\hfill \footnotesize 978-1-6654-8744-3/22/\$31.00  \copyright 2022 IEEE\hfill}
}
\makeatother

\begin{document}

\title{Artificial Pupil Dilation for Data Augmentation in Iris Semantic Segmentation\\
\thanks{This research work has been partially funded by the TOC Biometrics R\&D center SR-226 and SOVOS.}
}

\author{
\IEEEauthorblockN{Daniel P. Benalcazar}
\IEEEauthorblockA{\textit{R\&D Center SR-226} \\
\textit{TOC Biometrics}\\
Santiago, Chile \\
daniel.benalcazar@tocbiometrics.com}
\and

\IEEEauthorblockN{David A. Benalcazar}
\IEEEauthorblockA{\textit{Pontificia Universidad} \\
\textit{Catolica del Ecuador}\\
Quito, Ecuador \\
dbenalcazar234@puce.edu.ec}
\and

\IEEEauthorblockN{Andres Valenzuela}
\IEEEauthorblockA{\textit{R\&D Center SR-226} \\
\textit{TOC Biometrics}\\
Santiago, Chile \\
andres.valenzuela@tocbiometrics.com}

}


\maketitle


\begin{abstract}
Biometrics is the science of identifying an individual based on their intrinsic anatomical or behavioural characteristics, such as fingerprints, face, iris, gait, and voice. Iris recognition is one of the most successful methods because it exploits the rich texture of the human iris, which is unique even for twins and does not degrade with age. Modern approaches to iris recognition utilize deep learning to segment the valid portion of the iris from the rest of the eye, so it can then be encoded, stored and compared. This paper aims to improve the accuracy of iris semantic segmentation systems by introducing a novel data augmentation technique. Our method can transform an iris image with a certain dilation level into any desired dilation level, thus augmenting the variability and number of training examples from a small dataset. The proposed method is fast and does not require training. The results indicate that our data augmentation method can improve segmentation accuracy up to 15\% for images with high pupil dilation, which creates a more reliable iris recognition pipeline, even under extreme dilation.    
\end{abstract}

\renewcommand\IEEEkeywordsname{Keywords}


\section{Introduction}
In modern times, it is ever more frequent to use biometric systems daily from unlocking smartphones to border controls. Those systems rely on the distinctive features of each individual, such as the shape of their face, the patterns of fingerprints, veins and iris tissue, and even the micro-expressions when one talks or walks \cite{sundararajan2018deep}. Iris recognition in particular analyzes the pattern of the human iris that is made out of both changes in melanin coloration and structural folds that occur when this tissue changes the aperture of the pupil \cite{bowyer2008image, Hosseini2010}. The advantages of using the iris as a biometric pattern are that there is a high variability between individuals, the iris is protected by the cornea so it is unlikely to degrade over time, and it is a non-contactive, non-invasive and non-intrusive method \cite{daugman2009iris, bowyer2016handbook}. That is why India, for example, is using iris, along with fingerprints, for its Aadhaar national identification system of more than a billion individuals \cite{Daugman2014}.  

\begin{figure}[t!]
    \begin{center}
        \subfloat[Original image\label{subfig:Original_Image}]{{\includegraphics[width=0.32\linewidth]{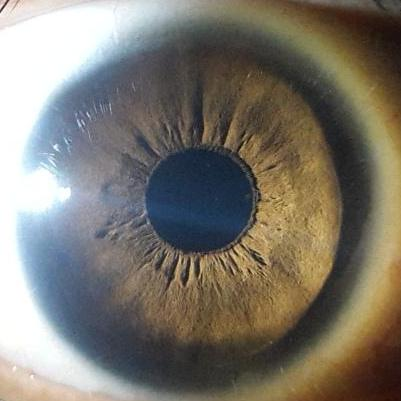} }}%
        \subfloat[Reduced dilation \label{subfig:Reduced_Dilation}]{{\includegraphics[width=0.32\linewidth]{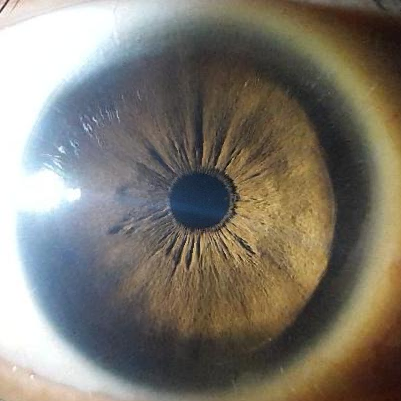} }}
        \subfloat[Increased dilation \label{subfig:Increased_Dilation}]{{\includegraphics[width=0.32\linewidth]{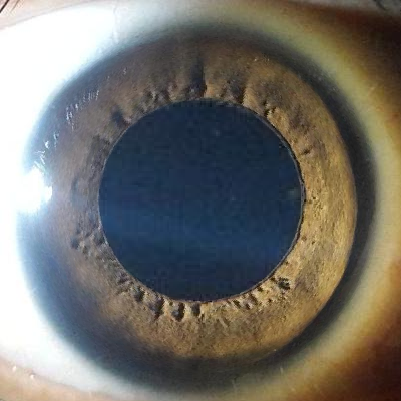} }}
    \end{center}
    \caption{Example of artificial dilation change. a) Original input image, dilation level $\lambda = R_{pupil}/R_{iris} = 0.282$. b) Artificially decreased dilation $\lambda = 0.150$. c) Artificially increased dilation $\lambda = 0.550$.}%
    \label{fig:dilation}%
\end{figure}

\begin{figure}[t]
    \begin{center}
        \subfloat[Iris Image\label{subfig:Iris}]{{\includegraphics[width=0.32\linewidth]{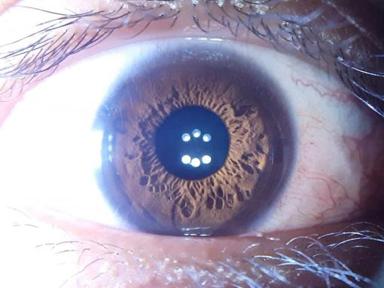} }}%
        \subfloat[Segmentation Mask \label{subfig:Mask}]{{\includegraphics[width=0.32\linewidth]{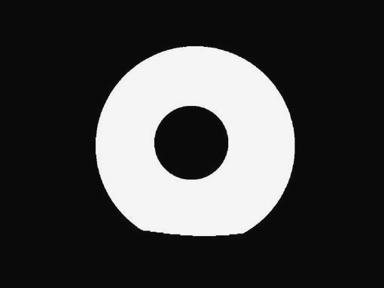} }}
        \subfloat[Localization \label{subfig:Loc}]{{\includegraphics[width=0.32\linewidth]{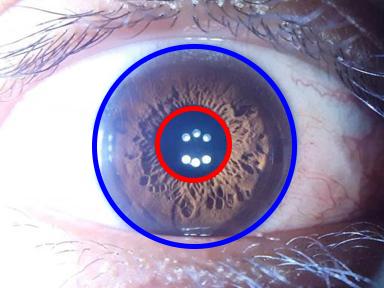} }}
    \end{center}
    \begin{center}
        \subfloat[Normalized Rubber Sheet\label{subfig:rubb}]{{\includegraphics[width=0.98\linewidth]{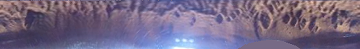} }}%
    \end{center}
    \caption{Iris segmentation and normalization. a) Input image. b) Segmentation mask produced by deep-learning methods. c) Iris and pupil localization. d) Normalized iris image, known as rubber sheet.
    }
    \label{fig:segmentation}
\end{figure}

Our method, illustrated in Fig.~\ref{fig:dilation}, aims to improve robustness in iris segmentation. The process of Iris recognition, shown in Fig.~\ref{fig:segmentation}, relies on the following steps. Iris segmentation, where the binary mask that encompasses the pixels of the iris is found (Fig.~\ref{subfig:Mask}) \cite{daugman2009iris}. Iris localization, in which the coordinates of the circles or ellipses that best fit the pupil and iris are computed (Fig.~\ref{subfig:Loc}) \cite{bowyer2016handbook}. Normalization, where the annular iris surface is turned into a rectangle known as the Rubber Sheet (Fig.~\ref{subfig:rubb}) \cite{daugman2009iris}. Encoding, where the information of the iris surface is encoded in a compact feature vector, which in some cases is binary \cite{bowyer2016handbook}. Finally, comparison, where the encodings of two irises are compared to find if they belong to the same subject or not. This step can also compare one iris against a dataset to determine the identity of the subject \cite{bowyer2016handbook}.

In this research work, we focus on improving the first link of the chain, the segmentation stage, by making it robust for a wide range of pupil dilations. The dilation level $\lambda$ is defined as the ratio between the pupil radius and the iris radius \cite{Hollingsworth2009}. Traditional iris datasets were captured under controlled illumination environments, and small dilation variation \cite{bowyer2016handbook}. However, studies have shown that images with extreme pupil dilation, in which the pupil is either too narrow or too large, can degrade iris recognition performance \cite{Hollingsworth2009, arora2012iris, Tomeo-Reyes2016}. Those extreme cases are not uncommon since the iris can dilate because of illumination changes, as well as alcohol and drug consumption \cite{arora2012iris, Tomeo-Reyes2016}. Therefore, it is essential to obtain datasets with representative cases of those examples for modern deep-learning algorithms to generalize and produce good segmentation results at any dilation level \cite{tapia2021semantic}.

That is why we propose a Data-Augmentation (DA) algorithm that artificially changes the dilation of input iris images, as seen in Fig.~\ref{fig:dilation}. In this way, the segmentation network's training process benefits from a wider variety of dilated images. This technique, in turn, makes the network more robust to extreme dilation conditions, as in the case of alcohol consumption \cite{tapia2021semantic}. We expect our method to increase the segmentation accuracy of any segmentation network, so we trained and tested seven of the most commonly used segmentation networks with and without our technique and compared the results.

As for the contributions of this work, we produced:
\begin{itemize}
\item An open-source DA function that artificially dilates the pupil in segmentation masks and iris images captured under Visible Light (VL) and Near Infrared Range (NIR) illumination.
\item A fast DA method that increases iris segmentation performance in extremely dilated images, without requiring a training process. 
\item A benchmarking evaluation, showing the improvement in segmentation performances on state-of-the-art networks, as well as its the T-test significance.
\end{itemize}


\section{Related Methods}

Original Iris Recognition methods relied on deterministic image processing techniques \cite{bowyer2008image, Daugman2014, bowyer2016handbook}. However, in recent years deep-learning methods have obtained a better performance since they can generalize well for wider populations and less restricted environments \cite{tapia2021semantic, gangwar2016deepirisnet, zhao2017towards, nguyen2017iris, minaee2019deepiris, zhao2019deep, mishra2019ccnet, yiu2019deepvog, lei2022attention}. Some do not require training in the recognition stage \cite{nguyen2017iris, zambrano2022notrain}, and others even make use of the 3D structure of the iris tissue \cite{benalcazar2020a, benalcazar2020b}. 

A problematic facing all those methods is the need for extensive labeled datasets to train the segmentation network, \cite{tapia2021semantic, mishra2019ccnet, yiu2019deepvog}. Most available massive datasets, such as CASIA, UBIRIS and Notre Dame, were captured under controlled illumination conditions and specific capture devices \cite{OMELINA2021datasets}. On the other hand, newer, more challenging datasets are captured in the wild under a huge variety of conditions; however, they either contain fewer images or are not publicly available \cite{tapia2021semantic, OMELINA2021datasets}. The greater the number of examples with more variability, the better deep-learning methods can perform in real-world situations \cite{tapia2021semantic, mishra2019ccnet, fang2020open}.  

Hollingsworth et. al \cite{Hollingsworth2009}, and Arora et. al \cite{arora2012iris} studied how less restricted illumination conditions, as well as alcohol consumption, during image capture would detriment recognition performance. This is because, in extreme dilation conditions, the iris is more difficult to segment and the available iris area is too small to extract the biometric pattern successfully \cite{Hollingsworth2009, arora2012iris}. Some research works have studied mathematical models for pupil dilation in order to mitigate this issue \cite{Tomeo-Reyes2015, Tomeo-Reyes2016}, while Tapia et. al \cite{tapia2021semantic} captured an extensive dataset of irises in normal conditions, as well as under the influence of alcohol in order to train better segmentation networks capable of operating even under extreme dilation or constrictions.  

An alternative to manually capturing and pre-processing a large dataset is to apply DA techniques on a small sample to increase the richness of the data artificially. Most common DA techniques include image translation, rotation, flipping, affine transformations, brightness and contrast adjustment, color shifting, and noise addition \cite{tapia2021semantic}. Most recently, Generative Adversarial Networks (GAN) have been used to artificially generate new iris images of nonexistent subjects, thus increasing the number of examples and enhancing performance \cite{lee2019conditional, tapia2019soft, maureira2021synthetic, maureira2022analysis}.
However, The variability of GAN-produced samples is limited by that of the original set \cite{chu2017cyclegan}; therefore, if few images contain extreme dilations, the generated images would too \cite{chu2017cyclegan}. To the best of our knowledge, there are no available methods that address the DA of the pupil dilation level from existing images. The proposed approach aims to solve that by generating new examples of iris images of the same individual but under any desired dilation level. In this way, new datasets with a uniform distribution of dilation levels $\lambda$ can be easily generated from readily available datasets.  

\section{Artificial Dilation Augmentation}
The proposed method is a DA function that can change the dilation level $\lambda_1$ of an iris image, or semantically segmented mask, into any desired dilation level $\lambda_2$. This function is based on traditional image processing and sampling methods, instead of deep learning, so it is fast and does not require any training.

In iris recognition, the coordinates of the circles that best fit the pupil and iris (Fig.~\ref{subfig:Loc}) are used to normalize the image, by means of polar coordinates. In this process, the annular region of the iris surface is transformed into a rectangle (Fig.~\ref{subfig:rubb}), known as rubber sheet, using \eqref{eq:normalization} \cite{daugman2009iris}. This normalization is not affected by dilation changes since the space between the pupil and iris is always normalized in the radial axis between 0 and 1 \cite{daugman2009iris}. In other words, all of the iris images of a subject at any dilation level $\lambda_1, \lambda_2, \lambda_3, ... \lambda_n$ will be transformed into the same rubber sheet.
Inspired by this normalization function, we sought an answer to the following question. Can the inverse of this transformation be used to synthesize an image with any desired dilation level  $\lambda_i$ from the rubber sheet? 

\begin{equation}
I(x(r,\theta),y(r,\theta)) \rightarrow I(r,\theta)
\label{eq:normalization}
\end{equation}

\begin{figure}[t]
\centering
\includegraphics[width=1\linewidth]{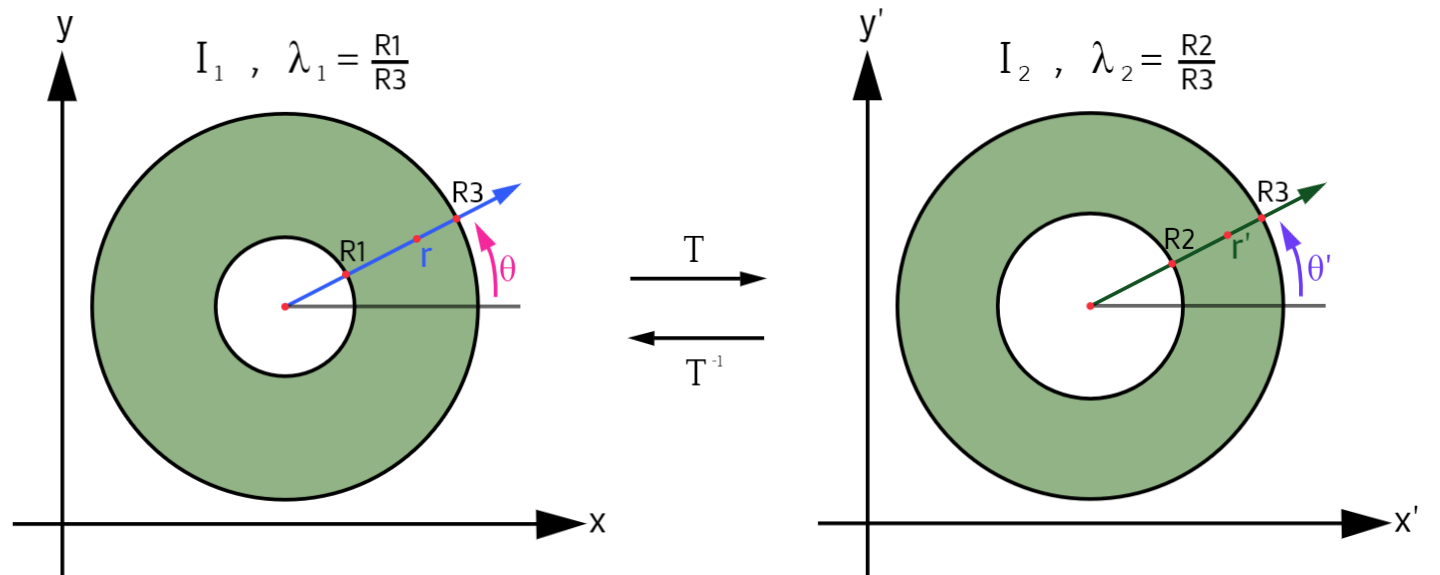}
\caption{Dilation Transformation Diagram.}
\label{fig:transform}
\end{figure}

By developing \eqref{eq:normalization}, we were able to derive the equation that can directly sample the coordinates of the image with dilation level $\lambda_1$ to produce a new image with dilation level $\lambda_2$, without even going to the normalized state. The resulting sampling equations can be best explained with the diagram of Fig.~\ref{fig:transform}. 
Let $I_1$ and $I_2$ be two iris images at dilation states $\lambda_1$ and $\lambda_2$ respectively, $R_1$ and $R_2$ be the radius of the pupil iris boundary of these images, and  $R_3$ be the radius of the iris sclera boundary in both images. The polar coordinates of a point in the iris surface of $I_1$ are represented by $(r,\theta)$, and the coordinates of the corresponding point in $I_2$ are given by $(r',\theta')$. The transformation $I_1 \rightarrow I_2$ is noted as $T$, and $T^{-1}$ is the transformation $I_2 \rightarrow I_1$. Then, we can start to deduce the relationships between the coordinates of the two images.

We can first note that the angle is not going to be affected by this transformation. Otherwise, $I_2$ would be rotated. Therefore, we can conclude that, $\theta$ is the same as $\theta'$. Then, only the radial component is going to be affected by the transformation. Since the normalization process is linear \cite{daugman2009iris}, we can conclude that, the distance between $R_1$ and  $r$ would be proportional to the  distance between $R_2$ and  $r'$ as presented in \eqref{eq:proportion}.

\begin{equation}
\frac{r-R_1}{R_3-R_1} = \frac{r'-R_2}{R_3-R_2}
\label{eq:proportion}
\end{equation}

Then we can solve for $r$ in order to know which point in $I_1$ must be sampled for a given $r'$ in the output image $I_2$ \eqref{eq:rIris}.

\begin{equation}
r = \frac{R_3-R_1}{R_3-R_2}(r' - R_2)+R_1 
\label{eq:rIris}
\end{equation}

Equation \eqref{eq:rIris} is valid for points in the iris surface; however, for the points inside the pupil, a slight variation, given in \eqref{eq:prop2} and \eqref{eq:rPupil}, must be used since they start when $r=0$.

\begin{equation}
\frac{r-0}{R_1-0} = \frac{r'-0}{R_2-0}
\label{eq:prop2}
\end{equation}

\begin{equation}
r= \frac{R_1}{R_2} r' 
\label{eq:rPupil}
\end{equation}

For points outside the iris, $r$ is the same as $r'$ since points on the sclera and the skin are not affected by dilation. Equation \eqref{eq:Transf} summarizes transformation $T^{-1}$ for the entire image.

\begin{equation}
T^{-1}:
\begin{cases}
    r_{r',\theta'} = (R_1 / R_2) r'
    & \text{,  if  } r' < R_2 \\
    r_{r',\theta'} = m  (r'-R_2)+R_1 
    & \text{,  if  } R_2 \leq r' < R_3 \\
    r_{r',\theta'} = r'
    & \text{,  if  } R_3 \leq r' \\
    \theta_{r',\theta'} = \theta'
    & \text{,  if  } 0 \leq \theta' < 2 \pi 
\end{cases}
\label{eq:Transf}
\end{equation}

In \eqref{eq:Transf}, the constant value $m$ is equal to $(R_3-R_1)\div(R_3-R_2)$. Finally, the algebraic expressions in \eqref{eq:sin_cos} can be used to transform from cartesian coordinates to polar coordinates, and thus $T^{-1}$ can be used on the square grid of pixels of an image. 

\begin{equation}
x = r \cos(\theta) \quad , \quad y = r \sin(\theta)
\label{eq:sin_cos}
\end{equation}

To synthesize image $I_2$ from $I_1$, the intensity value of every pixel in $I_2$ comes from sampling the corresponding pixel in $I_1$, whose coordinates are given by solving \eqref{eq:Transf} and \eqref{eq:sin_cos}. Pixels are sampled with nearest-neighbor interpolation because it is the fastest sampling technique \cite{corke2011robotics}. This approach is useful to reduce DA time while training semantic segmentation networks. However, other sampling techniques such as bilinear or bicubic can easily be implemented \cite{corke2011robotics}.

\begin{figure}[t]
    \begin{center}
        \subfloat[Original images \label{subfig:original}]{{\includegraphics[width=0.31\linewidth]{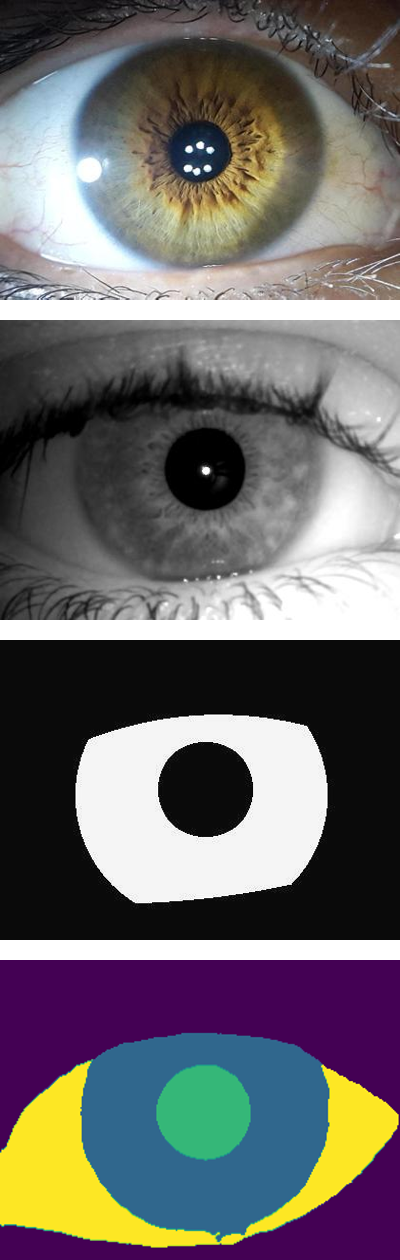} }}
        \subfloat[Dilated $\lambda = 0.45$ \label{subfig:dil1}]{{\includegraphics[width=0.31\linewidth]{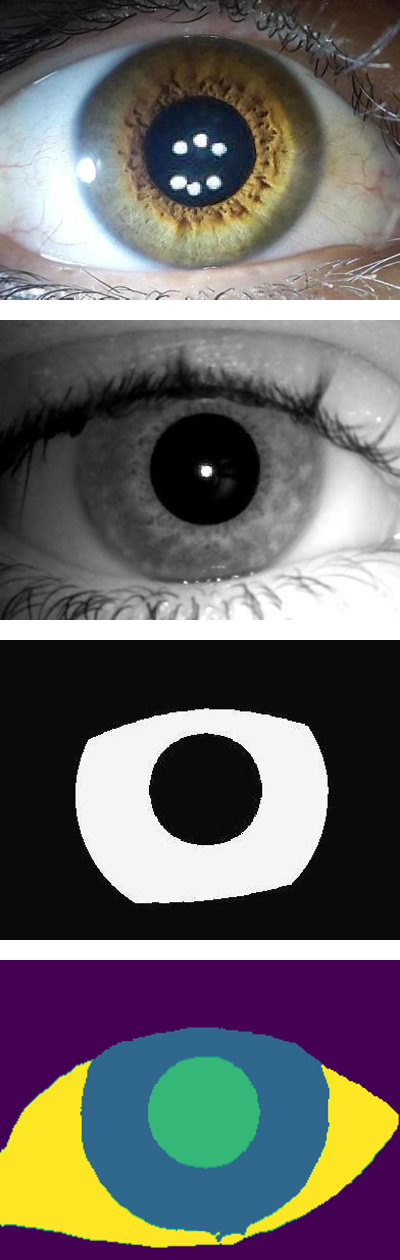} }}
        \subfloat[Dilated $\lambda = 0.6$
        \label{subfig:dil2}]{{\includegraphics[width=0.31\linewidth]{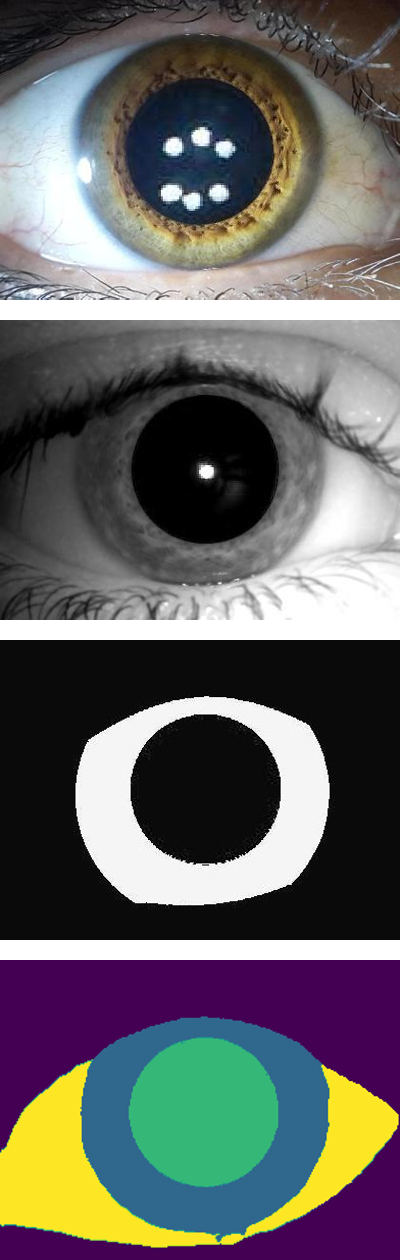} }}
    \end{center}
    \caption{Examples of artificially changing the dilation of a diversity of iris images using the proposed method. From top to bottom: Visible Light RGB image, Near Infrared Range grayscale image, binary mask, semantic mask with four classes. }
    \label{fig:examples}
\end{figure}

This method was applied for grayscale images, RGB images, binary segmentation masks, as well as multichannel semantic masks, as seen in Fig.~\ref{fig:examples}. The only artifact is that eye leads that cross the iris are affected by $T^{-1}$, as if the subject opened their eyes. Nevertheless, the output images look very realistic, which is essential for network generalization.

The proposed method, implemented both in Python and Matlab, is publicly available on GitHub\footnote {https://github.com/dpbenalcazar/ArtificialDilation}. It contains the raw DA functions, as well as didactic Graphical User Interface (GUI) examples.


\section{Datasets}
\label{section:datasets}
In this work we trained and tested several readily available semantic-segmentation networks using the datasets described in this section, in order to evaluate the performance of the proposed method. We selected two datasets, captured with the same capture device. These datasets were chosen to help testing the generalization capabilities of the networks in the presence of a wide range of pupil dilations.

\begin{figure}[t!]
    \centering
    \includegraphics[width=0.85\linewidth]{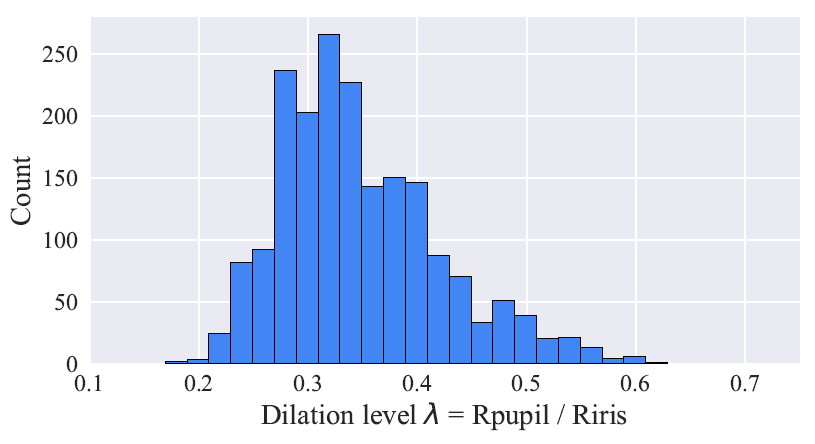}
    \caption{Train Dataset Dilation Distribution}
    \label{fig:train_hist}
    
    \vspace{5mm}
    
    \centering
    \includegraphics[width=0.85\linewidth]{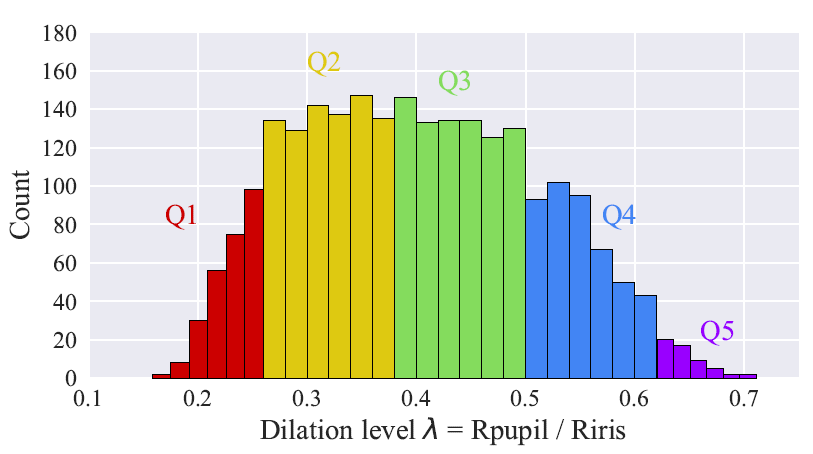}
    \caption{Test Dataset Dilation Distribution by Quantiles}
    \label{fig:test_hist}
\end{figure}

For training and validation, we utilized the dataset captured by Benalcazar et al. \cite{Benalcazar2019}, which tested iris recognition performance under two kinds of illumination. The commonly used NIR frontal illumination against frontal and lateral VL. This dataset consists of 96 subjects, five images of the right eye per subject, and 4 cases of illumination. There are 1,920 images in total, which were partitioned, in a person-disjoint manner, as 1,540 images for training and 380 for validation. The illumination conditions were controlled, so there is a small pupil dilation variation from image to image. The dilation distribution for the 1,920 images, seen in Fig.~\ref{fig:train_hist}, has the following parameters: mean=0.347, STD=0.073, min=0.169, max=0.625.   

For testing, we used the dataset that Benalcazar et al. \cite{benalcazar2020a} originally captured to train a network that predicts a 3D model of the iris tissue. For a better generalization, the authors captured a dataset with a wide range of pupil dilation. The subjects were exposed to darkness for 10 seconds, which dilated their pupils, then lights were turned on, and the contraction of the pupil was captured frame by frame for 3 seconds \cite{benalcazar2020a}. In that experiment, 26,520 images were captured; however, in this work, we utilized only a portion by uniformly sampling the dilation of each subject. Therefore, the test set in this work containes 120 subjects and ten images per eye of each subject, for a total of 2,400 images. The dilation distribution in this case, shown in Fig.~\ref{fig:test_hist}, has the following parameters: mean=0.4, STD=0.112, min=0.158, max=0.711. It can be noted that both the standard deviation and the max dilation are greater in the test set; thus, learning meaningful features from the training set is very challenging. That is why the chosen datasets are perfect to evaluate the effects of the proposed DA method. 

\begin{table*}[!t]
\caption{Mean IoU ± Standard Deviation for the Entire Test Set (2,400 Images)}
\centering
\begin{tabular}{c c c c c c c c }
\hline
\hline
  \textbf{Augmentation} &	\textbf{CCNet} &	\textbf{DenseNet10} &	\textbf{Vgg16-Unet} &	\textbf{ResNet50-Unet} & 	\textbf{MobileNet2-Unet} &	\textbf{pix2pix} & \textbf{CycleGAN} \\
\hline
  Normal &	0.681 ± 0.113 &	0.870 ± 0.068 &	0.854 ± 0.057 &	0.876 ± 0.060 &	0.857 ± 0.051 & 0.812 ± 0.116 &	0.741 ± 0.082 \\
  Artif. Dilation	& 0.722 ± 0.141 &	0.879 ± 0.057	& 0.875 ± 0.058	& 0.880 ± 0.053	& 0.880 ± 0.050	& 0.894 ± 0.055	& 0.750 ± 0.089 \\
\hline
  Improvement &	5.89\% &	1.07\%	& 2.43\% &	0.45\% &	2.67\%	& \textbf{10.09\%} &	1.25\% \\
\hline
\hline
\end{tabular}
\label{tab:1}
\end{table*}

\begin{table*}[!t]
\caption{Percentage of Mean IoU Improvement for each Quantile of the Test Set.}
\centering
\begin{tabular}{c c c c c c c c c c }
\hline
\hline
  \textbf{Quantile} &
  \textbf{Mean $\lambda$} &	\textbf{CCNet} &	\textbf{DenseNet10} &	\textbf{Vgg16-Unet} &	\textbf{ResNet50-Unet} & 	\textbf{MobileNet2-Unet} &	\textbf{pix2pix} & \textbf{CycleGAN} &	\textbf{Mean} \\
\hline
  Q1 &	0.2378 &	-0.33\%	& 0.79\%	& 1.61\%	& 0.98\%	& 1.49\% &	-0.02\%	& 0.54\% &	\textbf{0.72\%} \\
  Q2 &	0.3250 & 2.97\% &	1.22\% &	2.01\%	& 0.21\%	& 2.33\% &	2.14\% &	0.35\% & \textbf{1.60\%} \\
  Q3 &	0.4327 &	6.49\%	& 0.72\% &	2.16\%	& 0.26\%	& 2.83\%	& 9.91\% &	0.57\% &	\textbf{3.28\%} \\
  Q4 &	0.5361 &	12.30\% &	1.71\% &	3.40\% &	0.45\% &	3.43\%	& 28.36\% &	2.54\% &	\textbf{7.46\%} \\
  Q5 &	0.6307 &	22.58\% &	0.30\% &	6.56\%	& 2.08\% &	4.96\% &	56.80\% &	12.90\% &	\textbf{15.17\%} \\
\hline
\hline
\end{tabular}
\label{tab:2}
\end{table*}


\section{Experiments}

The purpose of our method is to help any iris semantic segmenter to perform better, specially under extreme dilation cases. In order to test the efficacy of our DA algorithm, we trained and tested seven readily available networks, with the previously described datasets, in two scenarios: Normal DA, and  Artificial Dilation DA. Both scenarios were trained from scratch with the same hyper-parameters per network.

\subsection{Data Augmentation Scenarios}
\begin{figure}[t!]
    \begin{center}
        \subfloat[Input image\label{subfig:input}]{{\includegraphics[width=0.32\linewidth]{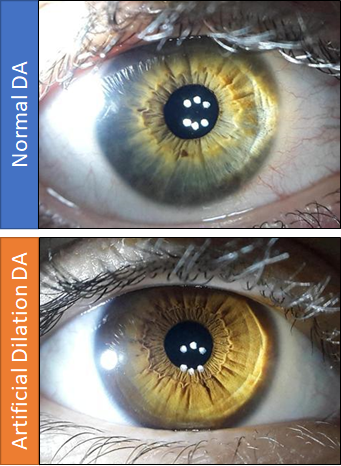} }}%
        \subfloat[Offline \label{subfig:ofline}]{{\includegraphics[width=0.32\linewidth]{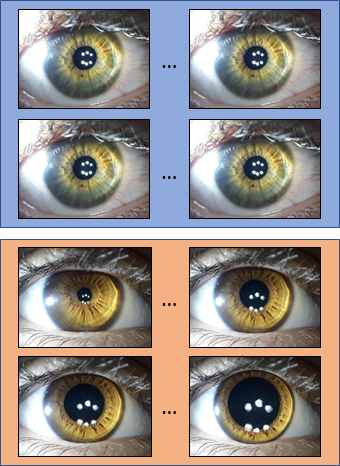} }}
        \subfloat[Online \label{subfig:online}]{{\includegraphics[width=0.32\linewidth]{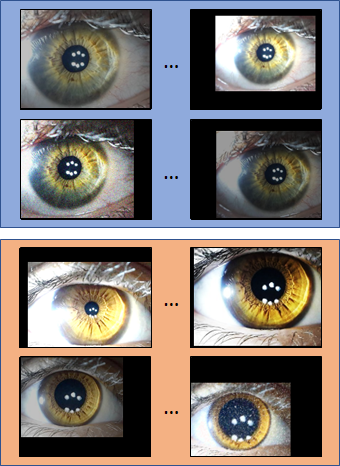} }}
    \end{center}
    \caption{DA schemes used in the experiments. a) Input image. b) Twenty offline augmentations. c) Random online DA during network training.}%
    \label{fig:augmentation}
\end{figure}

Figure~\ref{fig:augmentation} illustrates the DA applied in the two scenarios. For Normal DA, the offline augmentation consisted of making 20 copies of the input images. Therefore, the networks were presented with 38,400 images each epoch. Then, during training, a random transformation of translation, scaling, flipping, brightness, and blooring was applied to each image.

In Artificial Dilation DA, for each input image, we generated 19 artificial dilations using $\lambda$ equally separated between 0.15 and 0.75. Adding the original input images, we also have 38,400 images each epoch. It is worth mentioning that we generated the augmented images offline for consistency purposes; however, the proposed method can be used online as well. After that, the same random transformations as Normal DA were applied online to each image while training the networks.

\subsection{Trained Networks}
The seven networks trained in this work are the following. CCNet \cite{mishra2019ccnet, fang2020open} and DenseNet10 \cite{tapia2021semantic}, which were conceived for the purpose of iris semantic segmentation. They are based on the encoder-decoder architecture \cite{tapia2021semantic}. Using the same architecture, we employed three general purpose networks that have Unet \cite{ronneberger2015unet} in the decoder, and VGG16 \cite{simonyan2014vgg}, ResNet50 \cite{he2016resnet} and MobileNet2 \cite{howard2017mobilenets} in the encoder respectively. Finally, we trained pix2pix and CycleGAN \cite{chu2017cyclegan}, which are GANs that can translate the image to the semantic mask space. The semantic ground-truth masks used for training and testing have only two labels: iris and background.

In the hardware level, these networks require Graphics Processing Units (GPU) to accelerate training; however, during inference, they can produce results in a fraction of a second, even with single board computers such as the Raspberry Pi \cite{fang2020open}. The networks were trained on a Ryzen5 machine with 16GB of RAM and a 12GB GPU.

\subsection{Performance Metric}
The metric that we use for performance evaluation is the bitwise Intersection over Union (IoU) \cite{tapia2021semantic}. This metric, described in \eqref{eq:IoU}, compares the ground truth and predicted segmentation masks ($M_A$ and $M_B$) by counting the number of bits that are high in a logical AND operation between both masks, divided by the count of logical ones in the logical OR operation between them. 

\begin{equation}
IoU=\frac{\sum_{}^{} (M_A \wedge M_B)}{\sum_{}^{}  (M_A \vee M_B) +  \epsilon}
\label{eq:IoU}
\end{equation}

This metric takes the range of values between 0 and 1, being 1 a perfect match. The constant $\epsilon$ has a small positive value that ensures the denominator never takes a value of zero.


\section{Results}

The results of the described experiments, evaluated on the 2,400 test images, are presented in Table~\ref{tab:1}. This table shows the mean and STD values of the IoU metric, for each of the seven networks trained, with and without the proposed Artificial Dilation DA method. This table shows an increase in mean IoU value for all the trained models. The final row shows the relative improvement as a percentage. It can be seen that pix2pix and CCNet are the networks that improved the most with the proposed method. For instance, pix2pix increased mean IoU by 10.09\% (from 0.812 to 0.894). On the other side of the spectrum, DenseNet10 and ResNet50-Unet had a marginal improvement. This is because some networks have a richer architecture than others, thus, some can generalize well with normal DA, while others required the better examples provided by our method to learn. Overall, the average improvement produced by all the seven networks was 3.41\%. The computed p-value for the T-test applied on Table~\ref{tab:1} was 0.0394. This means that the increase in performance produced by our method was significant, with a probability of being caused by mere chance less than 5\%. This implies that the proposed DA method is a meaningful tool to increase performance in iris semantic segmentation.

Table~\ref{tab:2} shows the results divided by quantiles. This table analyzes the improvement margin when we divide the test set in five groups of equally separated dilation levels as depicted in Fig.~\ref{fig:test_hist}. The column Mean $\lambda$ shows the mean dilation level of each quantile. It can be seen that for Q1, Q2 and Q3, the improvement was less than 3.5\%. However, mean IoU improved by 7.46\% and 15.17\% in Q4 and Q5 respectively. This indicates that our method has a greater impact in improving segmentation results on iris images with high dilation levels.

Additionally, we tested the speed of the python implementation. We augmented 100 NIR images of $320 \times 280$ pixels. The average time per image was 12.014, 2.725 and 2.696 ms on a Ryzen3, Ryzen5 and Intel Core-I7 processors respectively. All machines had 16GB of RAM and used a single thread. The fastest time is equivalent to synthesizing 371 images per second, which is more than enough for real time training of a network.


\section{Conclusions}
We developed a DA function that artificially changes the dilation level $\lambda_1$ of an iris image into any desired level $\lambda_2$. This function enriches the dilation variability of any dataset, and can be applied both offline and online because of its speed. This method is based on sampling techniques, so it does not require any training.
The proposed method improved the overall IoU in semantic segmentation networks by 3.41\% in average, with a p-value of 0.0394. This means that our method significantly increased the performance of the trained networks.
We observed that the greatest improvement occurred in the quantile of highest dilation. In average, IoU increased by 15.17\%. This means that the proposed DA method helps achieving better performance in iris segmentation tasks, specially for dilated images, which are scarce in most available datasets. This in turn would lead to a better iris recognition pipeline.

Future work includes implementing more robust sampling methods, such as bilinear, and improving the equations so they work over non-concentric elliptic segmentation. This would expand the potential of this method for working with in-the-wild datasets, which consist of a limited number of iris images from different perspectives.


\section*{Acknowledgment}

This work was partially supported by TOC Biometrics R\&D center SR-226, and SOVOS. 
Spacial thanks to the Department of Electrical Engineering, Universidad de Chile.

\bibliographystyle{IEEEtran}
\bibliography{References.bib}

\end{document}